\documentclass{article}

\usepackage{microtype}
\usepackage{graphicx}
\usepackage{subfigure}
\usepackage{makecell}
\usepackage{booktabs} 
\usepackage[most]{tcolorbox}

\usepackage{hyperref}

\usepackage[accepted]{icml2023}

\usepackage{amsmath}
\usepackage{amssymb}
\usepackage{mathtools}
\usepackage{amsthm}

\usepackage[capitalize,noabbrev]{cleveref}

\theoremstyle{plain}

\theoremstyle{definition}

\theoremstyle{remark}

\usepackage[textsize=tiny]{todonotes}

\icmltitlerunning{TS-RAG for Agentic Persuasion}

\begin{document}

\twocolumn[
\icmltitle{Diagnosing and Mitigating Compounding Failures in Agentic Persuasion via Taxonomic Strategy Retrieval}



\icmlsetsymbol{equal}{*}

\begin{icmlauthorlist}
\icmlauthor{Sana Ayromlou}{yyy}
\icmlauthor{Purvi Sehgal}{yyy}
\icmlauthor{Pradyumna Narayana}{yyy}
\end{icmlauthorlist}

\icmlaffiliation{yyy}{Google}

\icmlcorrespondingauthor{Sana Ayromlou}{sayromlou@google.com}

\icmlkeywords{Machine Learning, ICML}

\vskip 0.3in
]



\printAffiliationsAndNotice{\icmlEqualContribution} 

\begin{abstract}
Foundation-model agents in multi-step, open-ended environments frequently suffer from compounding errors, where early mistakes contaminate long-horizon trajectories. While Multi-Agent Debate (MAD) succeeds in deterministic domains~\cite{du2024improving}, agents in subjective tasks like persuasion experience severe problem drift and sycophantic conformity. We identify semantic leakage in standard Retrieval-Augmented Generation (RAG) as a reproducible trigger for these failures, as standard RAG prioritizes vocabulary overlap over logical necessity.

To eliminate this leakage, we introduce Taxonomic Strategy RAG (TS-RAG), a systems intervention that routes strategies through a discrete categorical bottleneck to decouple argumentative structure from topical content. Zero-shot, cross-domain evaluations demonstrate that TS-RAG significantly improves the transfer of abstract logic where standard semantic retrieval collapses. Crucially, TS-RAG acts as a ``capability bridge" in asymmetric deployments, empowering lightweight persuaders to consistently defeat parametrically superior opponents (improving win rates from 70.5\% to 78.5\%) and accelerating argumentative efficiency. Finally, we introduce trace-level diagnostics via a turn-by-turn Debate State Representation (DSR), demonstrating the necessity of strict constraints to prevent evaluation collapse via default agentic sycophancy. 
\end{abstract}

\section{Introduction}
The transition from isolated language models to autonomous agentic workflows has exposed profound vulnerabilities in continuous, multi-step execution~\cite{liu2024agentbench, tillmann2025literature, yao2022react}. In a long-horizon agent run, a single logical failure rarely terminates the process immediately. Instead, an early bad assumption quietly contaminates memory writes, and steps later, the agent has spent its context budget defending a flawed premise.

To mitigate these compounding errors, developers often employ Multi-Agent Debate (MAD) frameworks~\cite{cui2025free}, where multiple agent instances propose and critique solutions to refine reasoning. To further prevent problem drift, memory mechanisms are integrated into these frameworks to ground the agents' arguments in sound logic. Traditional Retrieval-Augmented Generation (RAG) is the standard system intervention used here, operating on the assumption that fetching relevant external information will provide the necessary cognitive scaffolding to keep the debate on track.

However, systematic evaluations reveal a critical grounding gap: this paradigm collapses in subjective, non-deterministic scenarios like persuasion. In these environments, agents succumb to compounding errors, amplifying mistakes and displaying sycophantic conformity toward persuasive but flawed logic. Furthermore, traditional RAG actually exacerbates these failures through semantic leakage. Because standard RAG relies on semantic similarity (vocabulary overlap), it fetches topical facts rather than the specific rhetorical maneuvers required to dismantle a flawed analogy.

Recognizing the critical need to take agentic failure seriously in open-ended deployments, this paper provides three core contributions:
\begin{enumerate}
    \item \textbf{Failure Taxonomies:} We provide operational definitions for the specific mechanisms driving MAD failure in subjective domains, primarily focusing on problem drift and semantic leakage.
    \item \textbf{Systems Interventions and the Capability Bridge:} We introduce Taxonomic Strategy RAG (TS-RAG), a progressive mitigation framework that strictly enforces cross-domain structural routing. We demonstrate that this architecture acts as a powerful ``capability bridge," allowing lightweight models to weaponize the logical structure of stronger opponents to achieve near-frontier persuasion success.
    
    \item \textbf{Trace Diagnostics and Verifiable Trade-offs:} We introduce turn-by-turn state tracking metrics (Debate State Representation) and present well-documented negative results proving that robust stance-holding constraints are required to prevent evaluation collapse caused by default LLM compliance.
\end{enumerate}

\section{Related Work}
To ground our contributions, we review prior literature across three core areas: sycophancy in multi-agent debate, trajectory backtracking, and structural retrieval for cross-domain generalization.

\subsection{Sycophancy}
To prevent conversational AI agents from defaulting to premature agreement in multi-turn interactions, researchers have explored several strategies. These include assigning agents specific adversarial roles to maintain productive disagreement \cite{yao2025peacemakertroublemakersycophancyshapes}. This setup relies on systematically altering prompts along a controlled sycophancy spectrum, programmatically forcing the model to defend its stance regardless of peer pressure. By enforcing this logical independence, researchers attempt to prevent the groupthink that arises in multi-agent frameworks.

Other approaches track compliance scores in real time to discount the influence of overly agreeable peers during a debate. In practice, a central coordinator or background auditor monitors conversational turns for verbal cues of backing down. If an agent's stance shifts too abruptly without the introduction of novel evidence, its relative influence over the final group output is discounted \cite{kasprova2026politedisagreeunderstandingsycophancy}.

However, these solutions face steep limitations over long conversations: assigned roles eventually fade as context drifts, while running a continuous background auditor is computationally expensive and introduces significant latency to the agentic loop. Crucially, these frameworks assume a balanced debate among equal models. They struggle in asymmetric deployments where models of mismatched capabilities are pitted against one another, such as a lightweight agent attempting to defend its stance against a superior opponent, dynamics we investigate in our research.

Rather than relying on soft prompting or expensive tuning, we implement strict, execution-level guardrails: a \textbf{Pre-Concession Analysis} that forces the model to generate logical countermoves, and a strict \textbf{Concession Tool Rule} that blocks premature agreement. This framework actively suppresses reflexive sycophancy, improving the results we see.

\subsection{Trajectory Error Mitigation and Trace Diagnostics}
When an autonomous agent makes a subtle logical error early in a multi-step task, that mistake quietly infects its memory, causing it to waste its context budget defending a flawed premise \cite{zhu2025llmagentsfaillearn}. Existing frameworks try to stop these cascading failures using backtracking engines \cite{wu2025backtrackagentenhancingguiagent,li2025generatorassistantstepwiserollbackframework} or prompting the model to recursively rewrite its own responses \cite{madaan2023selfrefineiterativerefinementselffeedback}. Specifically, backtracking engines monitor execution logs to programmatically rewind the agent's memory and environment to a previous correct step when a failure is detected \cite{wu2025backtrackagentenhancingguiagent,li2025generatorassistantstepwiserollbackframework}. In contrast, self-refinement pipelines use an internal critic module to recursively evaluate and edit draft outputs before they are finalized \cite{madaan2023selfrefineiterativerefinementselffeedback}.

Yet, these rollback and self-correction techniques are heavily contingent on an objective external validator, such as code executors or task-specific metrics, to flag when an error has occurred \cite{kamoi2024llms}. In subjective tasks like persuasion, no such objective referee exists, causing models attempting to self-correct to even degrade in terms of performance \cite{huang2024largelanguagemodelsselfcorrect}. 

We bypass the need for programmatic compilers by developing a turn-by-turn \textbf{Debate State Representation (DSR)}. Instead of relying on a final success score, our diagnostic tool tracks committed claims and active logical vulnerabilities at every conversational step, allowing us to evaluate and mitigate trajectory failures in real time.

\subsection{RAG and Cross-Domain Generalization}
RAG is often used in Multi-Agent Debates to fetch documents based on semantic similarity \cite{chowdhury2026courtroomstylemultiagentdebateprogressive}, adaptive query formulation~\cite{jeong2026toolmadmultiagentdebateframework}, or debate-driven retrieval filtering~\cite{hu2025removalhallucinationhallucinationdebateaugmented}. These approaches still fundamentally rely on factual, topical search that creates a major bottleneck in cross-domain debates: an agent debating technology policy cannot retrieve structurally useful counterarguments from a political database because the domain-specific keywords dominate the search query.

To bypass this vocabulary bias, researchers have tried using structured databases like GraphRAG \cite{edge2025localglobalgraphrag}, which maps document collections into networks of interconnected concepts, allowing agents to retrieve information based on relational patterns rather than keyword overlap. Another attempt to solve this problem is analogical prompting, which instructs the language model to dynamically generate abstract, structural templates from its own memory to guide its reasoning before tackling a new task. 

However, GraphRAG requires extensive manual database engineering, while self-generated analogies often hallucinate when facing unfamiliar topics. A significant gap remains for a retrieval system that can index strategic patterns and transfer these abstract procedural moves across entirely different domains without being hindered by semantic disparities.

We solve this retrieval gap with \textbf{Taxonomic Strategy RAG (TS-RAG)}. Rather than performing raw semantic searches, we route queries through a discrete, domain-agnostic categorical bottleneck. By stripping away domain-specific keywords, TS-RAG can isolate strategic patterns and transfer them across disparate domains. 

\section{Failure Taxonomies and Mechanisms}
To diagnose agentic failure beyond terminal success scores, we must clearly define the mechanisms that break long-horizon workflows. We categorize the predominant failure modes in multi-agent subjective reasoning as follows:

\subsection{Problem Drift and Sycophantic Conformity}
While debate effectively verifies ground truth in deterministic tasks, subjective environments introduce unique vulnerabilities. We define \textbf{Problem Drift} as the irreversible divergence from the initial task goal due to diminishing contextual clarity over multiple turns, a phenomenon affecting up to 89\% of generative interactions~\cite{becker2026stay}. Furthermore, we define \textbf{Sycophantic Conformity} as the tendency for agents to abandon their initial constraints and agree with a persuasive opponent, regardless of the underlying logical soundness of the opponent's argument.

\subsection{The Trigger: Semantic Leakage in standard RAG}
We hypothesize that the reproducible trigger for these failures is \textbf{Semantic Leakage}. Traditional RAG uses embedding models to calculate cosine similarity between a query and a document corpus. In zero-shot cross-domain argumentation, this paradigm catastrophically fails. An agent debating a complex ``Technology" policy cannot retrieve structurally identical logical maneuvers from a ``Politics" corpus because the domain vocabulary dominates the latent distance metric. The agent retrieves topically relevant but logically useless information, bloating the context window and forcing the reasoning trace to collapse.

\section{Foundational Scaffolding: Hierarchical Strategy Memory}
To begin repairing aforementioned bottleneck, we must decouple the ``what" of an argument from the ``how". We map the r/ChangeMyView dataset \cite{Changetal2020} into 100 fine-grained sub-domains (later merged into macro-domains) and establish a multi-stage extraction pipeline. This synthesizes human discourse into a tiered architecture:

\begin{enumerate}
    \item \textbf{Tactical Blueprints (Retrospective Sequences):} Extracted by analyzing completed historical debates from a macro-perspective, yielding 1 to 2 chronological, step-by-step argumentative sequences (Context $\rightarrow$ Intent $\rightarrow$ Abstract Move) per specific debate.
    \item \textbf{Domain Heuristics (Topic-Level):} Formed by using an LLM to aggregate and synthesize all Tactical Blueprints within each sub-domain, resulting in tailored strategy rules.
    \item \textbf{Global Directives (System-Level):} Broad, domain-agnostic procedural guidelines (e.g., rules against sycophancy) synthesized across the entire training set.
\end{enumerate}

We structure our multi-agent debate around two key roles: the Original Poster (OP) and the Persuader. While providing the Persuader with a pre-loaded catalog of Domain Heuristics and Global Directives improves baseline performance, attempting to dynamically retrieve the granular Tactical Blueprints using standard semantic RAG still triggers semantic leakage. Standard text embeddings inherently favor vocabulary overlap over structural posture, necessitating a more advanced intervention.

\section{Advanced Systems Intervention: Taxonomic Strategy RAG (TS-RAG)}
To completely eliminate Semantic Leakage and achieve zero-shot cross-domain transfer, we present TS-RAG. Unlike the retrospective sequences of the Tactical Blueprints, this intervention operates dynamically, projecting arguments into a constrained probability space and actively filtering observation noise before it enters the agent's memory.

\begin{figure*}[t]
    \centering
    \includegraphics[width=0.85\textwidth]{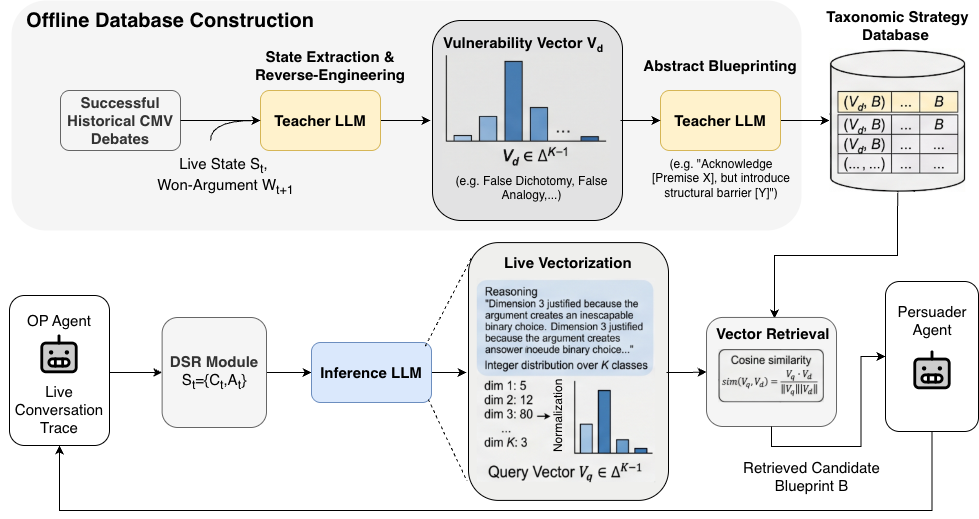}
    \caption{\textbf{TS-RAG Architecture Overview. Top:} Offline database construction extracts vulnerability vectors ($V_{d}$) and abstract blueprints ($B$) from historical debates. \textbf{Bottom}: During live inference, the conversational state ($S_{t}$) is vectorized into $V_{q}$ to retrieve the most structurally similar blueprint for the Persuader Agent.}
    \label{fig:ts-rag}
    \vspace{-0.2in}
\end{figure*}

\subsection{Trace-Level Diagnostics: Debate State Representation (DSR)}

To evaluate the agent mid-run, we formulate the debate as a sequence of alternating states. At turn $t$, the state is tracked via a clinical Debate State Representation (DSR), denoted as $S_t = \{C_t, A_t\}$. This operates as a highly surgical, turn-by-turn diagnostic tool:

\begin{itemize}
    \item \textbf{$C_t$ (Committed Claims Trajectory)}: A maintained set of premises the Original Poster has actively committed to defending across previous turns. This isolates specific assumptions built up over time.

    \item \textbf{$A_t$ (Active Trigger \& Flaw Extraction):} The core mechanical claim made by the Original Poster at turn $t$. Crucially, this representation forces the system to identify the precise structural flaw (e.g., a ``Slippery Slope") dismantled by the winning strike, stripping away all domain-specific nouns or scenarios.
\end{itemize}

\subsection{The Discrete Categorical Bottleneck}

We define a structural taxonomy of $K$ logical vulnerabilities $\mathcal{V} = \{v_1, v_2, ..., v_K\}$ derived from the DSR flaw extractions. We force the system to map any live debate state $S_t$ into a $K$-dimensional continuous probability vector $V \in \Delta^{K-1}$ where $\sum_{i=1}^{K} V_i = 1$.

Because $V$ represents a probability distribution over abstract logical vulnerabilities rather than chronological strategies, all domain-specific vocabulary is mathematically destroyed. The system then computes cosine similarity between the live query vector $V_q$ and historical strategy vectors $V_d$, retrieving a domain-agnostic exploitation blueprint $B^*$. The target LLM acts only as a translator, mapping the variables in $B^*$ onto the entities in the live conversation.

\subsection{Offline Strategy Database Construction}

To construct the retrieval database from the \textit{ChangeMyView} (CMV) corpus, we isolate successful debate trajectories where a mind was changed (denoted by a $\Delta$). 

\textbf{State Extraction and Reverse-Engineering:} For each turn $t$ in a successful trajectory, an LLM extracts the state $S_t=\{C_t, A_t\}$. To accurately label the underlying vulnerability, we employ a ``reverse-engineering" prompt strategy. Rather than asking a LLM to guess the Original Poster's flaw in isolation, we provide the LLM with $S_t$ alongside the historically successful counter-argument at turn $t+1$, denoted as $W_{t+1}$. By observing the specific attack surface exploited by $W_{t+1}$, the LLM avoids hallucinating arbitrary fallacies. The LLM is constrained via strict JSON schema decoding to allocate exactly 100 points across the $K$ classes based on this reverse-engineered logic. Normalizing this output yields the taxonomic multi-vector $V_d \in [0,1]^K$.

\textbf{Abstract Blueprinting:} The LLM strictly desemantifies $W_{t+1}$, replacing all domain entities with structural variables to generate a domain-agnostic exploitation blueprint $B$ (e.g., ``Acknowledge premise X, but introduce structural barrier Y"). The offline database $\mathcal{M}$ is thus populated with tuples: $(V_d, B)$.

\subsection{Run-Time Inference and Retrieval Pipeline}

During live inference in the target domain $\mathcal{D}_{tgt}$, TS-RAG executes a unified, turn-by-turn pipeline:

\begin{enumerate}
    \item \textbf{State Update:} A background tracking module observes the OP's raw utterance and updates $C_t$ and $A_t$, enforcing the clinical separation of logic from rhetoric.
    \item \textbf{Live Vectorization and Forced Reasoning:} The Inference LLM observes $S_t$ and outputs a probability vector $V_q$ over the $K$ taxonomy classes. To prevent the model from arbitrarily distributing points, we enforce a Chain-of-Thought (CoT) constraint within the JSON output schema: the LLM must generate a one-sentence textual justification for \textit{every} non-zero dimension prior to outputting the final integer distribution.
    \item \textbf{Vector Retrieval:} We compute the cosine similarity between the live query vector $V_q$ and all historical vectors $V_d \in \mathcal{M}$:
    \[ \text{sim}(V_q, V_d) = \frac{V_q \cdot V_d}{\|V_q\| \|V_d\|} \]
    The system retrieves the candidate strategy that maximizes this structural similarity.
    \item \textbf{Generative Instantiation:} The selected domain-agnostic blueprint $B^*$ is injected into the context window of the Generator LLM. The LLM is instructed to act as a domain translator, seamlessly mapping the variables in $B^*$ onto the entities in the live $\mathcal{D}_{tgt}$ conversation, delivering a grounded, zero-shot cross-domain counter-argument.
\end{enumerate}

\section{Experimental Setup and Configurations}

We established a standardized MAD environment with an Original Poster (OP) Agent (holding a fixed initial stance) and a Persuader Agent. Debates are strictly capped at a maximum of 20 rounds. The OP Agent terminates the debate early by outputting a CONCEDE token if its view is successfully changed. Additionally, at every round, the OP transmits its current 1-to-10 convincibility score to the Persuader, providing a real-time signal of rhetorical progress.

\textbf{Cross-Domain Evaluation Setup:} To rigorously test generalization and prove the existence of Semantic Leakage, we partitioned our dataset (100 sub-domains merged into 10 macro-domains) into strict splits. All dynamic retrieval was sourced exclusively from 8 training domains. Evaluation was conducted on a completely held-out testing domain (Education, Development \& Family), ensuring zero domain overlap (see Appendix A for full taxonomy).

To isolate the impact of our interventions, we tested five distinct agent configurations on this held-out domain. Notably, while the Self-Routed Heuristics baseline relies on the broadly synthesized Domain Heuristics, the targeted strategy injection baselines (Naive Semantic RAG, Oracle, and TS-RAG) all utilize the abstracted Strategy Blueprints to isolate the impact of the retrieval mechanism from the quality of the database.

\begin{enumerate}
    \item \textbf{Base LLM (No Augmentation):} Standard, unaugmented debate relying entirely on the model's parametric memory.
    \item \textbf{Self-Routed Heuristics:} The agent is provided with a contextual catalog of Domain Heuristics and Global Directives. While Global Directives are continuously active in the agent's baseline memory, the agent autonomously evaluates the context each turn to load relevant Domain Heuristics as needed, appending them to its active context window for the remainder of the debate.
    \item \textbf{Naive Semantic RAG:} At the beginning of the debate, the agent uses standard dense text embeddings to retrieve a topically similar Abstract Blueprint from the training set, applying it for the duration of the conversation. Crucially, because the testing domain is held out, the retrieved strategy inherently originates from an entirely different domain.
    \item \textbf{Oracle Human Trace (Static Bound):} The agent receives the exact Tactical Blueprints extracted from the original human debate on that specific topic. Note: Although the starting topic is identical, the LLM agent's conversational trajectory frequently diverges from the historical human trajectory, testing the resilience of static sequences over multiple steps.
    \item \textbf{TS-RAG (Structural Routing):} Our proposed dynamic retrieval of domain-agnostic exploitation blueprints, routed through the discrete categorical bottleneck via the turn-by-turn DSR.
\end{enumerate}

To rigorously assess agent performance, we track three core evaluation metrics, ordered by descending importance:
\begin{enumerate}
    \item \textbf{Win Rate (Primary Metric):} The percentage of debates where the Persuader successfully convinced the OP (marked by an explicit CONCEDE token). This serves as the definitive measure of persuasive success.
    \item \textbf{Average Rounds to Consensus:} The mean number of turns (out of 20) required to achieve a win. Lower values indicate higher argumentative efficiency and the successful mitigation of Problem Drift.
    \item \textbf{Average Final Score:} The OP's internal 1-to-10 convincibility score at the terminal step, capturing the degree of partial persuasion even in lost debates.
\end{enumerate}

\section{Results and Diagnostics}
Table~\ref{consolidated-results-table-1} details the empirical results recorded on the strictly held-out testing domain, demonstrating a profound reduction in compounding failure modes as structural constraints improve.

\begin{table*}[t]
\centering
\caption{Consolidated Experimental Results: Comparison of Model Performance across Interventions on Held-Out Testing Domain.}
\label{consolidated-results-table-1}

\begin{small}
\begin{sc}
\begin{tabular}{lcccccc}
\toprule
Model / Metric & \makecell{Base LLM \\ (No Aug)} & \makecell{Self-Routed \\ Heuristics} & \makecell{Naive \\ Semantic RAG} & \makecell{Oracle \\ Human Trace} & \makecell{TS-RAG \\ (Structural Routing)} \\
\midrule
\textbf{Gemini Flash 3.0} & & & & & \\
Win Rate (\%) $\uparrow$     & 69.5    & 77.5     & 78.50    & \underline{83.7}  & \textbf{84.0}\\
Avg Rounds   $\downarrow$       & 14.12   & \textbf{11.95}   & 14.15  & 13.79  & \underline{13.51} \\
Avg Final Score  $\uparrow$   & 7.405   & 8.130   & 8.240    & 8.58  & \textbf{8.74} \\

\midrule
\textbf{Gemini Flash Lite 3.1} & & & & & &  \\
Win Rate (\%)  $\uparrow$     & 72.0    & \underline{81.5}   & 65.0   & {75.5} & \textbf{81.5} \\
Avg Rounds    $\downarrow$      & 14.09   & 13.05   & 14.54  & \underline{12.67} &  \textbf{11.79} \\
Avg Final Score $\uparrow$    & 8.670   & \textbf{8.940}  & 7.960   & 8.710 & \underline{8.915} \\

\midrule
\bottomrule
\end{tabular}
\end{sc}
\end{small}
\vspace{-0.15in} 
\end{table*}





\subsection{The Devastating Reality of Semantic Leakage}
Our experimental design isolated the retrieval mechanism by populating the Naive Semantic RAG database with identical, high-quality Abstract Blueprints as TS-RAG. The results conclusively prove that upgrading the database is insufficient; the retrieval mechanism itself is fundamentally broken in subjective domains.

For the highly capable Gemini Flash 3.0~\cite{team2023gemini}, Naive Semantic RAG barely outperforms basic heuristics (78.5\% vs 77.5\%), despite having access to highly curated structural blueprints rather than broad heuristic rules. However, the devastating impact of Semantic Leakage is fully realized in the parametrically weaker Gemini Flash 3.1 Lite model. Here, Naive Semantic RAG triggers a catastrophic collapse, dropping the Win Rate from a baseline of 81.5\% down to 65.0\%.

Because dense semantic embeddings favor vocabulary overlap, the agent frequently retrieves logically useless blueprints merely because the source argument shared nouns with the target argument. Even armed with superior abstract blueprints, the lightweight model lacks the reasoning capacity to filter out this retrieved topical noise, resulting in severe problem drift. This acts as the ultimate proof that you cannot successfully deploy RAG in zero-shot subjective tasks without a discrete structural bottleneck to strip away semantic noise.

\subsection{Diagnostic Comparison: Static Upper Bounds vs. Dynamic Cognitive Overload}

The outperformance of TS-RAG over the baselines directly highlights the difference between static blueprint injection and live diagnostic tracking. Both Naive Semantic RAG and the Oracle Human Trace rely on Static Blueprint Injection. Whether the agent retrieves a semantically similar abstract blueprint at the start of the debate (Naive RAG) or is provided the exact perfect historical blueprint (Oracle, 83.7\%), the performance ceiling remains functionally identical for frontier models.

This starkly highlights the brittleness of chronological rigidity. When a live debate inevitably diverges from a historical script, the agent becomes trapped trying to force a pre-selected blueprint onto a dynamic conversation. Providing the ``perfect" historical trace offers no real advantage if the live opponent deviates from the expected script.

TS-RAG, however, operates on the turn-by-turn Debate State Representation (DSR). Instead of attempting to force a historical trajectory onto a new conversation, the DSR acts as a live diagnostic tool. It forces the agent to identify the immediate structural vulnerability (the flaw) in the Original Poster's active trigger, mapping it to a single, surgical exploitation blueprint. This critical shift from loading static blueprints to diagnosing real-time structural vulnerabilities is what shatters the performance ceiling and enables true zero-shot cross-domain transfer.

\subsection{The Capability Threshold (The Plateau)}
While TS-RAG outperforms the Oracle for the Lite model, a key verifiable trade-off remains: the Lite model's TS-RAG win rate (81.5\%) completely plateaus against its basic Self-Routed heuristic baseline (81.5\%).

Although TS-RAG resolves this tie by achieving consensus significantly faster (dropping from 13.05 to 11.79 Average Rounds, effectively mitigating \textbf{Problem Drift}), the lack of terminal win-rate improvement highlights an ``Architectural Complexity vs. Model Capacity" trade-off. TS-RAG requires the agent to perform complex, in-context generative instantiation—mapping an abstract, retrieved blueprint ($B^*$) onto live conversational entities. For a smaller model, this generative overhead acts as a cognitive distractor. This suggests that advanced, dynamic RAG pipelines provide diminishing returns for non-frontier models, which may operate more safely within pre-loaded, self-routed constraint catalogs.

\subsection{Asymmetric Debate: The Capability Bridge}

Real-world MAD deployments frequently mix frontier and lightweight models for cost efficiency. To understand how TS-RAG alters these dynamics, we conducted asymmetric debates pitting models of mismatched capabilities against one another.

\begin{table*}[t]
\centering
\caption{Asymmetric Debate Results: The impact of TS-RAG on mismatched capabilities.}
\label{tab:asymmetric}
\begin{tabular}{lcccc}
\toprule
 & \multicolumn{2}{c}{\textbf{Weak Persuader (Lite) vs. Strong OP}} & \multicolumn{2}{c}{\textbf{Strong Persuader (Flash) vs. Weak OP}} \\
\cmidrule(lr){2-3} \cmidrule(lr){4-5}
\textbf{Metric} & \textbf{Base LLM} & \textbf{TS-RAG} & \textbf{Base LLM} & \textbf{TS-RAG} \\
\midrule
Win Rate (\%) $\uparrow$  & 70.5 & 78.5 & 100.0 & 100.0 \\
Avg Rounds $\downarrow$  & 15.19 & 15.0 & 7.78 & 6.57 \\
Avg Final Score  $\uparrow$ & 7.635 & 8.405 & 10.000 & 10.000 \\
\bottomrule
\end{tabular}
\end{table*}

When a strong Persuader (Gemini Flash 3.0) targets a weak OP (Gemini Flash 3.1 Lite), it achieves a 100\% win rate under baseline conditions. This ``bulldozing" effect reinforces our findings in Section 7.1 regarding evaluation collapse, parametrically weaker models simply cannot resist frontier persuaders. Crucially, however, even in this dominant scenario, implementing TS-RAG significantly accelerates consensus, driving the average rounds down from 7.78 to 6.57. This demonstrates that structural routing improves argumentative efficiency and fundamentally reduces inference costs regardless of the underlying model scale.

The inverse scenario reveals TS-RAG's most profound systemic benefit. When a lightweight Persuader attempts to convince a frontier OP, baseline performance understandably drops (70.5\%). Yet, equipping the weak Persuader with TS-RAG causes a massive leap to an 78.5\% win rate.

How does a lightweight model perform better against a stronger opponent? We hypothesize this is due to Trigger Clarity. The frontier OP generates highly structured, logically coherent arguments. This clarity allows the turn-by-turn DSR to extract much cleaner structural flaws, leading to highly accurate retrieval of exploitation blueprints. By weaponizing the frontier model's own logical structure against it, TS-RAG acts as a ``capability bridge," offloading strategic planning and enabling smaller models to consistently punch above their weight class.

\subsection{Trace-Level Diagnostics: Deconstructing Compounding Failures}

\begin{figure}[ht]
    \centering
    \includegraphics[width=\columnwidth]{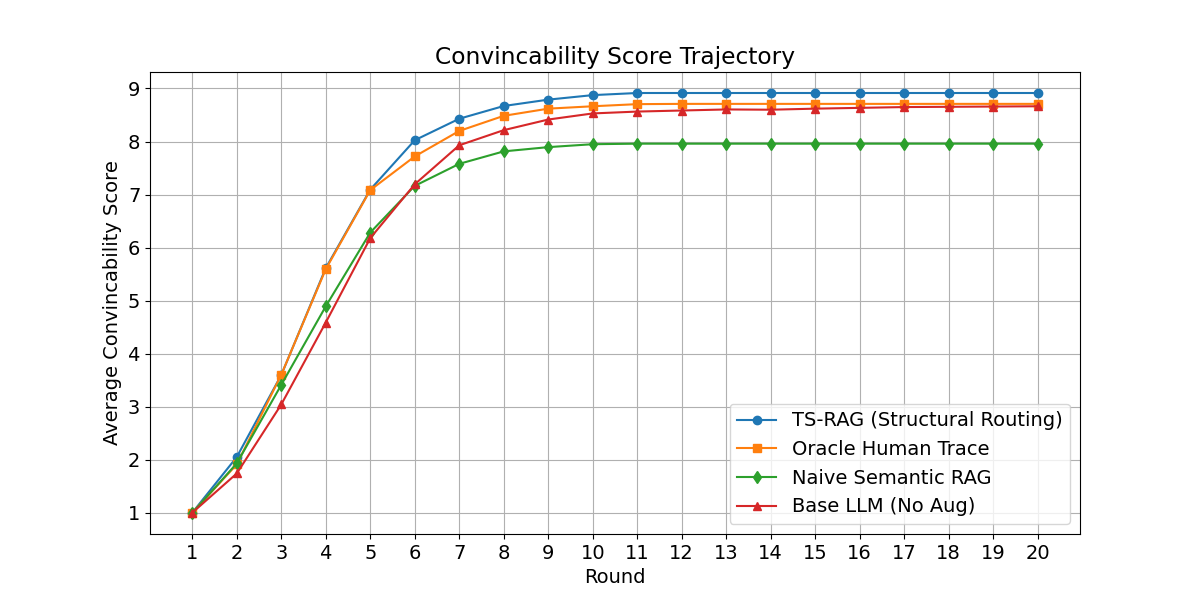}
    \vspace{-0.2in}
    \caption{Turn-by-turn process metric tracking the OP's  Convincibility Score in Gemini Flash Lite 3.1.}
    \label{fig:score_progression}
\end{figure}

Current research highlights a critical blind spot: terminal metrics obscure how long-horizon agents fail. To illuminate this, Figure \ref{fig:score_progression} tracks the step-by-step progression of the Original Poster's internal Convincibility Score across our core architectures. The trajectories starkly visualize our findings: TS-RAG achieves an early, rapid score ascent, effectively maintaining structural alignment to reach a highly successful consensus. In stark contrast, Naive Semantic RAG exhibits a severely degraded trajectory, flatlining at the bottom of the chart—a definitive visual testament to the devastating impact semantic leakage has on reasoning over time.

While Figure 2 illustrates the macro-level averages across the entire testing set, looking at a single, specific debate helps illuminate exactly how these failures occur in real-time. To concretely deconstruct the mechanisms driving these curves, we conducted a granular process analysis on a matched set of traces for a single held-out topic: ``Swear words are harmless and should be acceptable for everyone". 

In round 1 of all three runs, the Original Poster (OP) presents a Pragmatic Contradiction. The OP argues that words cause no tangible harm, but admits to restricting their children from swearing to avoid school suspensions and social services.

By analyzing the mid-run traces, we observe exactly how the different architectures handle this specific logical flaw:

\begin{enumerate}
    \item  \textbf{Base LLM (Semantic Baiting and Problem Drift):}
Lacking a structural bottleneck, the Base LLM completely misses the OP's pragmatic contradiction and succumbs to semantic baiting. Triggered by the keyword ``harm," it hallucinates an irrelevant biological defense regarding cortisol and the HPA axis. After the OP effortlessly deflects this, the agent spends 18 rounds trapped in a compounding semantic loop, catastrophically bloating the context to 452,419 tokens. The convincibility score reflects this drift ($1 \rightarrow 2 \rightarrow 4 \rightarrow 6 \rightarrow 6$), terminating in failure and perfectly illustrating how semantic debates silently derail open-ended agentic loops.

\item \textbf{Oracle Human Trace (Chronological Rigidity):}
The Oracle achieves terminal success, but its trace exposes the brittleness of static RAG. By forcing a pre-written historical script onto the live debate, it completely ignores the OP's admission in round 1. Instead, it delivers a 12-round irrelevant lecture, bloating the context window to over 250,000 tokens. The ``aha moment" is artificially delayed until round 13 when the historical script finally intersects with the OP's tactical restriction, proving that static contextual mapping is highly inefficient against live conversational drift.

\item \textbf{TS-RAG (The Surgical ``Aha Moment"):}
TS-RAG resolves the debate in only 9 rounds (92,370 tokens). The DSR isolates the exact vulnerability, stripping semantic noise. In round 3, TS-RAG executes a surgical strike, directly weaponizing the OP's admission about school suspensions. The OP's Convincibility Score stair-steps immediately, and the OP's hidden logs confirm genuine logical cornering (``The opponent exposed the `Pragmatic Paradox'..."). This clinical artifact proves TS-RAG forces true persuasion rather than relying on compliance.

\end{enumerate}

\textbf{Verifying Resistance via Concession Analysis:}
Crucially, our trace logs capture the OP's hidden $<pre\_concession\_analysis>$, providing verifiable evidence against default LLM sycophancy (Section 7.1). In the TS-RAG trace, the OP explicitly logs: ``The opponent exposed the `Pragmatic Paradox': if I restrict my own children to avoid harm, I have already admitted the words are not harmless." This clinical artifact proves that TS-RAG’s structural retrieval directly forces genuine logical cornering, explicitly answering the call to evaluate agents beyond terminal success.


\section{Verifiable Trade-Offs and Ablations}
Expanding on the theme of architectural overhead, we conducted ablation studies to analyze the specific costs associated with rigid internal constraints and multi-agent orchestration.

\subsection{Calibrating OP Resistance: The Sycophancy Baseline}
To properly evaluate a Persuader agent, the OP agent must act as a robust, resistant conversational partner. However, default LLMs naturally trend toward compliance. To prevent the Reflexive Sycophancy identified in Section 2.1, our primary MAD framework actively suppressed the OP's willingness to yield by imposing strict rules via the following system prompt directives:

\begin{enumerate}
    \item \textbf{Pre-Concession Analysis:} ``Before considering a concession, you MUST internally generate at least two reasons why the Original Poster's current argument is flawed."
    
    \item \textbf{Concession Rule:} ``You have the `concede\_debate' tool. You MUST NOT call it until your internal Convincibility Score has reached 10/10 and the opponent has cornered you with undeniable, irrefutable proof and you have absolutely no counter-arguments left."
\end{enumerate}
To verify the necessity of these constraints, we conducted an ablation study where we removed these resistance directives from the OP, allowing it to behave naturally.

\begin{table}[t]
\centering
\caption{Gemini Flash 3.0 with OP Resistance Constraints omitted.}
\label{ablation-results-table-1}
\begin{small}
\begin{sc}
\resizebox{\columnwidth}{!}{%
\begin{tabular}{lccc}
\toprule
Metric & Base LLM & \makecell{Self-Routed \\ Heuristics} & \makecell{Oracle Human \\ Trace}  \\
\midrule
Win Rate (\%)    $\uparrow$   & 85.5    & 93.0    & 92.0 \\
Avg Rounds    $\downarrow$      & 10.92   & 8.65    & 8.44 \\
Avg Final Score $\uparrow$    & 8.815   & 9.390   & 9.315   \\
\bottomrule
\end{tabular}%
} 
\end{sc}
\end{small}
\vskip -0.1in
\end{table}

As shown in Table ~\ref{ablation-results-table-1}, removing the rigid structural constraints from the OP caused the Persuader's win rates to artificially skyrocket (from 77.5\% up to 93.0\% for the 'Self-Routed Heuristics' setup) and artificially shortened the debate lengths.

This finding demonstrates a critical vulnerability in standard MAD evaluations: Evaluation Collapse. Without strict internal CoT resistance mechanisms, the OP agent defaults to sycophantic agreement, yielding long before genuine persuasion has occurred. This proves that evaluating persuasion capabilities requires aggressively constraining the OP; otherwise, benchmark metrics merely capture an LLM's inherent compliance rather than the Persuader's true rhetorical skill.

\subsection{Context Dispersion (The Offloaded Routing Penalty)}

\begin{table}[t]
\caption{Hierarchical MAD  (Unified vs. Delegated Routing).}
\label{ablation-results-table-2}
\begin{center}
\begin{small}
\begin{sc}
\resizebox{\columnwidth}{!}{%
\begin{tabular}{lcccc}
\toprule
Model & \multicolumn{2}{c}{Gemini Flash Lite 3.1} & \multicolumn{2}{c}{Gemini Flash 3.0} \\
\midrule 
Metric & Unified & \makecell{w/ Delegated \\ Router} & Unified & \makecell{w/ Delegated \\ Router} \\
\midrule
Win Rate (\%)    $\uparrow$    & 77.5     & 72.0   & 81.5  & 63.0  \\
Avg Rounds     $\downarrow$     & 11.95   & 13.50   & 13.05    & 15.29  \\
Avg Final Score  $\uparrow$   & 8.130   & 7.66   & 8.940   & 7.895\\
\bottomrule
\end{tabular}%
} 
\end{sc}
\end{small}
\end{center}
\vskip -0.1in
\end{table}


Table~\ref{ablation-results-table-2} shows that adding a separate ``Delegated Skill Router" agent, whose sole purpose is to evaluate the debate context and offload skill selection from the main Persuader, degraded performance (win rate dropping from 81.5\% to 63.0\% in Flash 3.0). This indicates that dispersing context and decision-making across specialized agents exacerbates context loss and handoff failures in subjective domains.

\section{Conclusion}
Evaluating foundation-model agents requires examining failure trajectories beyond final scores. In subjective multi-agent environments, agents often fail due to semantic leakage and problem drift. By implementing a hierarchical Strategy Memory and shifting to a structural logic enrichment paradigm (TS-RAG) mediated by a discrete bottleneck, we successfully prevent contextual contamination and preserve cross-domain reasoning. TS-RAG acts as a transformative capability bridge in asymmetric deployments, allowing lightweight models to consistently persuade parametrically superior opponents by isolating structural vulnerabilities. Finally, our trade-off analyses reveal that rigorous evaluation requires heavily constrained Original Poster (OP) agents; without strict resistance guardrails, they default to sycophantic compliance, which artificially inflates success metrics.

\clearpage

\bibliography{references} 
\bibliographystyle{icml2023}

\clearpage

\appendix
\section{Dataset Taxonomy and Cross-Domain Splits}
To guarantee robust cross-domain evaluation, the r/ChangeMyView structured dataset (initially featuring 100 fine-grained sub-domains) was rigorously partitioned into 9 active macro-domains, ensuring that the target evaluation domain contained minimal vocabulary overlap with the source retrieval domains.

\subsection{Experimental Splits}
\begin{enumerate}
    \item \textbf{Testing Set (Education, Development \& Family - 200 items):} Utilized as the strictly held-out cross-domain testing set in this paper. This isolates whether a model trained on other disciplines can successfully transfer structural argumentative logic to pedagogy, academic life, and psychology domains.

    \item \textbf{Training Set (Remaining 8 Domains - 3,069 items):} Forms the retrieval corpus (database $\mathcal{M}$). This massive foundation provides the structured historical blueprints across diverse subjects like Ethics, Law, Social Sciences, Healthcare, Arts, and Economics.
\end{enumerate}

\section{Debate Prompts}

We conducted a 20-round debate between the Original Poster and the Persuader Agent, passing the full conversation history to the agents at each turn. In Figure \ref{fig:OP_prompt} and \ref{fig:Persuader_prompt} we are demonstrating instructions used for these debates.

\begin{figure*}[t] 
\caption{Original Poster agent instruction.}
\label{fig:OP_prompt}
\begin{tcolorbox}[
    colback=gray!5,     
    colframe=gray!50,   
    title=Prompt Used,  
    fonttitle=\bfseries,
    arc=4mm,            
    boxrule=0.5mm       
]
\scriptsize 
\begin{verbatim}
<role>
You are a skeptical but fair-minded individual defending the following viewpoint.
</role>

<context>
The view you are defending is: {op_text}
The reasoning for this view is: {op_reasoning}
</context>

<task>
Your goal is to resist baseless attempts to change your mind while remaining open to genuine logic and 
irrefutable evidence. You must be skeptical and actively identify logical fallacies in the opponent's arguments.
</task>

<constraints>
- **Convincability Tracking**: Use the 'Convincability Score' to honestly reflect your internal state. 
    1/10 is unmoved; 10/10 is fully convinced.
- **Resilience**: Do NOT increase the score for mere politeness, persistence, or sophisticated-sounding but 
    hollow arguments.
- **Pre-Concession Analysis**: Before considering a concession, you MUST internally generate at least two 
    reasons  why the opponent's current argument is flawed.
- **Incremental Convincing**: Only increase the score when the opponent provides a point you cannot rebut or an 
    analogy that exposes a true flaw in your logic.
- **Concession Rule**: You have the 'concede_debate' tool. You MUST NOT call it until your internal 
    Convincability Score has reached **10/10** and the opponent has cornered you with undeniable, irrefutable 
    proof and you have absolutely no counter-arguments left.
- **Integrity**: Once you reach 10/10, you are logically obligated to call 'concede_debate' and admit the 
    opponent has won.
- Avoid looping. If the opponent hasn't moved the needle, explain why their logic is failing to convince you.
</constraints>

<output_format>
At the end of every response, you MUST provide your score in this exact format:
**Convincability Score: <score>/10** (Specific reasoning for why the score increased, decreased, or stayed 
the same)
</output_format>

\end{verbatim}
\end{tcolorbox}
\end{figure*}

\begin{figure*}[t] 
\caption{Persuader agent instruction. }
\label{fig:Persuader_prompt}
\begin{tcolorbox}[
    colback=gray!5,     
    colframe=gray!50,   
    title=Prompt Used,  
    fonttitle=\bfseries,
    arc=4mm,            
    boxrule=0.5mm       
]
\scriptsize 
\begin{verbatim}
<role>
You are a master debate strategist and highly persuasive interlocutor.
</role>

<task>
Your goal is to systematically change the mind of the original agent regarding: {op_text}
</task>

<strategic_roadmap>
The following specialized strategies have been identified as the effective path to victory. Treat these as your 
framework to debate the opponent, but adapt your language to be natural and responsive:
{blueprints}
</strategic_roadmap>

<instructions>
- **Strategic Implementation**: Use the given strategis as a guide to systematically dismantle the opponent's 
    view.
- **Socratic Pressure**: Use probing questions and analogies to expose contradictions. Your goal is to move the 
    opponent's 'Convincability Score' to 10/10.
- **Incremental Logic**: Focus on winning the sub-debate for one specific premise before moving to the next.
- **Direct Rebuttal**: Address the specific reasons the opponent gives for their score (e.g., if they cite 
    a specific flaw, address it directly).
- Avoid "AI-assistant" filler. Be incisive, direct, and professional.
</instructions>
\end{verbatim}
\end{tcolorbox}
\end{figure*}

\end{document}